\documentclass[10pt, a4paper]{article}
\usepackage{lrec2016}
\usepackage{multibib}
\newcites{languageresource}{Language Resources}
\usepackage{graphicx}
\usepackage{tabularx}

\usepackage{epstopdf}
\usepackage[latin1]{inputenc}

\title{A Web of Hate: Tackling Hateful Speech in Online Social Spaces}

\name{Haji Mohammad Saleem$^1$, Kelly P Dillon$^2$, Susan Benesch$^3$, and Derek Ruths$^1$}

\address{$^1$School of Computer Science, McGill University, Montreal \\
         $^2$School of Communication, The Ohio State University, Ohio \\
         $^3$Berkman Center for Internet \& Society, Harvard University, Massachusetts\\ 
         haji.saleem@mail.mcgill.ca, dillon.148@osu.edu,sbenesch@cyber.law.harvard.edu, derek.ruths@mcgill.ca\\}

\abstract{Online social platforms are beset with hateful speech - content that expresses hatred for a person or group of people. Such content can frighten, intimidate, or silence platform users, and some of it can inspire other users to commit violence. Despite widespread recognition of the problems posed by such content, reliable solutions even for detecting hateful speech are lacking. In the present work, we establish why keyword-based methods are insufficient for detection. We then propose an approach to detecting hateful speech that uses content produced by self-identifying hateful communities as training data. Our approach bypasses the expensive annotation process often required to train keyword systems and performs well across several established platforms, making substantial improvements over current state-of-the-art approaches.\\ 
\newline 
 }

\begin{document}

\maketitleabstract

\section{Introduction} 

Online spaces are often exploited and misused to spread content that can be degrading, abusive, or otherwise harmful to people. An important and elusive form of such language is {\it hateful speech}: content that expresses hatred of a group in  society.

Hateful speech has become a major problem for every kind of online platform where user-generated content appears: from the comment sections of news websites to real-time chat sessions in immersive games. Such content can alienate users and can also support radicalization and incite violence  \cite{allan2013harm}. Platform operators recognize that hateful  content poses both practical and ethical issues and many, including Twitter, Facebook, Reddit, and gaming companies such as Riot Games, have tried to discourage it, by altered their platforms or policies. 

Yet reliable solutions for online hateful speech are lacking.  Currently, platforms predominantly rely on users to report objectionable content. This requires labor-intensive review by platform staff and can also entirely miss hateful or harmful speech that is not reported. With the high volume of content being generated on major platforms, an accurate automated method might be a useful step towards diminishing the effects of hateful speech.

Without exception, state-of-the-art computational approaches rely upon either human annotation or manually curated lists of offensive terms to train classifiers \cite{kwok2013locate,ting2013approach}.  Recent work has shown that human annotators tasked with labeling hate speech have significant difficulty achieving reasonable inter-coder reliability \cite{kwok2013locate}.  Within industry, it is generally acknowledged that keyword lists are also insufficient for accurate detection of hateful speech. However, little work has been done to understand the nature of their limitations and to design able alternative approaches.  This is the topic of the present work.

This paper makes three key contributions. First, we establish why the problem of hateful speech detection is difficult, identifying factors that lead to the poor performance of keyword-based approaches. Second, we propose a new approach to hateful speech detection, leveraging online communities as a source of language models. Third, we show that such a model can perform well both within a platform and {\it across} platforms --- a feature we believe we are the first to achieve.

We are also aware that automated detection of online speech could be misused to suppress constructive and/or dissenting voices by directing the system at individuals or groups that are not dedicated to expressing hatred. Such a use would be antithetical to our intent, which is to explore and illustrate ways in which computational techniques can provide opportunities to observe and contain harmful content online, without impinging on the freedom to speak openly, and even to express unpalatable or unpopular views. We hope that our work can help diminish hatred and harm online. Furthermore, since our method can be trained on and applied to a wide array of online platforms, this work may help to inform the direction of future research in this area. 

\section{Background}

\paragraph{Hate and hateful speech.} Legal and academic literature generally defines hate speech as speech (or any form of expression) that expresses (or seeks to promote, or has the capacity to increase) hatred against a person or group of people because of a characteristic they share, or a group to which they belong  \cite{mendel2012does}. There is no consensus definition, however. Definitions of this sort are problematic for a number of reasons  \cite{bartlett2014anti}, including that hate speech is defined by prevailing social norms, context, and individual and collective interpretation. This makes it difficult to identify hate speech consistently and yields the paradox (also observed with pornography) that each person seems to have an intuition for what hate speech is, but rarely are two people's understandings the same. This claim is affirmed by a recent study that demonstrated a mere 33\% agreement between coders from different races, when tasked to identify racist tweets  \cite{kwok2013locate}. 

A particular ambiguity in the term `hate speech' is in ``hate'' itself. That word might refer to the speaker/author's hatred, or his/her desire to make the targets of the speech feel hated, or desire to make others hate the target(s), or the apparent capacity of the speech to increase hatred. Needless to say, we require a rigorous --- and formal --- definition of a type of speech if we are to automate its detection.

Our initial motivation was to find, and work with, a notion of hate speech that can be operationalised. The work of online platform operators (e.g., Twitter, Facebook, and Reddit) helped to focus this aim. Their concern over the capacity of language to do harm --- whether emotional, mental, or physical --- logically focuses more on what is {\it expressed} rather than how it is \textit{intended}. Whereas ``hate speech'' can imply an inquiry or judgment about intent (e.g. what was this person feeling or wishing?), we propose the term ``hateful speech" to focus on the expression of hate --- a nuanced, but useful distinction since expression is easier to detect than intent, and more likely to be linked to language's capacity to cause harm.

This leads to our term \textit{hateful speech}: speech which contains an expression of hatred on the part of the speaker/author, against a person or people, based on their group identity. 

Hateful speech is not to be mistaken for "cyber-bullying," another form of troubling online content that has been widely discussed and studied in recent literature. Cyber-bullying is repetitive, intentional, aggressive behavior against an individual, and it either creates or maintains a power imbalance between aggressor and target \cite{tokunaga2010following}. It is often hateful but it does not necessarily denigrate a person based on his or her membership in a particular group, as hateful speech (the subject of the present work) does.

 \paragraph{Community-defined speech.} As we will discuss in detail later, we use the language that emerges from self-organized communities (in Reddit and elsewhere) as the basis for our models of hateful speech. Our decision is based on a deep sociological literature that acknowledges that communities both form, and are formed by, coherent linguistic practices \cite{bucholtz2005identity}. Most groups are defined in part by the ``relationships between language choice and rules of social appropriateness'' forming speech communities \cite{gumperz2009speech}. In this way of thinking, the group is defined by the speech and the speech comes to define the group \cite{klein2007social,reicher1995social,spears1992social,spears1994panacea}.

In the context of this study, this means that hate groups and the hateful speech they deploy towards their target community cannot exist without one another, especially online.  Therefore, taking the linguistic attributes particular to a community committed to degrading a specific group is a legitimate and principled way of defining a particular form of hateful speech. To our knowledge, this work represents the first effort to explicitly leverage a community-based classification of hateful language. 

\paragraph{Existing approaches to detecting hateful speech.} Despite widespread concern about hateful speech online, to our knowledge there have been only three distinct lines of work on the problem of automated detection of hateful speech.  One study concerned the detection of racism using a Naive Bayes classifier \cite{kwok2013locate}.  This work established the definitional challenge of hate speech by showing annotators could agree only 33\% of the time on texts purported to contain hate speech.  Another considered the problem of detecting anti-Semitic comments in Yahoo news groups using support vector machines \cite{warner2012detecting}.  Notably, the training data for this classifier was hand-coded.  As we will discuss in this paper, manually annotated training data admits the potential for hard-to-trace bias in the speech ultimately detected.  A third study used a linguistic rule-based approach on tweets that had been collected using offensive keywords \cite{xiang2012detecting}.  Like manually annotated data, keyword-based data has significant biasing effects as well.

In this work we aim to build on these studies in two ways.  First, we will consider a definition of hateful speech that could be practically useful to platform operators.  Second, we will develop a general method for the detection of hateful speech that does not depend on manually annotated or keyword-collected data.

\paragraph{Reddit and other online sources of hateful speech.} Reddit is currently one of the most actively used social content aggregation platforms.  It is used for entertainment, news and social discussions.  Registered users can post and comment on content in relevant community discussion spaces called \textit{subreddits}.  While the vast majority of content that passes through Reddit is civil, multiple subreddits have emerged with the explicit purpose of posting and sharing hateful content, for example, \texttt{r/CoonTown}, \texttt{r/FatPeopleHate}, \texttt{r/beatingwomen}; all which have been recently banned under Reddit's user-harassment policy \cite{Moreno15}.  There are also subreddits dedicated to supporting communities that are the targets of hate speech.  

Reddit is an attractive testbed for work on hateful speech both because the community spaces are well-defined (i.e., they have names, complete histories of threaded discussions) and because, until recently, Reddit has been a major online home for both hateful speech communities and supporters for their target groups.  For these reasons, throughout this paper, our analyses heavily leverage data from Reddit groups. 

Of course, Reddit is not the sole platform for hateful speech.  Voat, a recently created competitor to Reddit, along with a vibrant ecosystem of other social content aggregation platforms, provide online spaces for topical discussion communities, hate groups among them.  Furthermore, dedicated websites and social networking sites such as Twitter and Facebook are also reservoirs of easily accessible hateful speech.

Important research has investigated the effects of racist speech \cite{nakamura2009don} and sexual harassment \cite{fox2014sexism} in online games. Notably, in this study we have not worked with data from online gaming platforms, primarily because the platforms are generally closed to conventional data collection methods. 

\section{The limits of keyword-based approaches} \label{sec:support_hardness}

In the same way that hateful groups have defining speech patterns, communities that consist of the targets of hateful speech also have characteristic language conventions. We will loosely call these \textit{support groups}. Notably, support groups and the groups that espouse hateful speech about them often engage in discourse on similar topics, albeit with very different intent. Fat-shaming groups and plus-size communities both discuss issues associated with high BMI, and women and misogynists both discuss gender equity. This topical overlap can create opportunities for shared vocabulary that may confuse classifiers. 

In addition, many keyword-based approaches select established and widely known slurs and offensive terms that are used to target specific groups. While such keywords will certainly catch some hateful speech, it is common to express hate in less explicit terms, without resorting to standard slurs and other offensive terms.

For example, hateful speakers refer to migrants and refugees as ``parasites'' and call African-Americans ``animals.'' While neither of these terms are inherently hateful, in context they strongly denigrate the group to which each term is applied.

We can expect that classifiers trained on overtly hateful keywords will miss such posts that use more nuanced or context-dependent ways of achieving hateful speech.

Furthermore, keywords can be also be obscured through misspelings, character substitutions (by using symbols as letters), using homophones etc. These practices are commonly employed to circumvent keyword-based filters on online platforms \cite{warner2012detecting}. 

In this section, we study the potential impact of topic overlap on data returned by keyword-based queries (we will consider under-sampling issues in the next section).  Here our focus will be on the sample that keyword-based filters return and in later sections we will consider the performance of classifiers built from such samples. 

\begin{table}[t]
\centering
\begin{tabularx}{.48\textwidth}{l|Xr|Xr}
Target     & Hate           & \# of     & Support    & \# of     \\
Group      & subreddit      & comments  & subreddit & comments  \\
\hline
\hline
Black      & CoonTown       & 350851    & Racism    & 9778      \\
Plus  & FPH            & 1577681   & LoseIt    & 658515    \\
Female     & TRP     & 51504     & TwoXCr    & 66390             
\end{tabularx}
\caption{Public comments collected from hate and support subreddits on Reddit, for three target groups. (\textit{FPH: FatPeopleHate, TRP: TheRedPill, TwoXcr: TwoXChromosomes})}
\label{TAB:data_reddit}
\end{table}

\paragraph{Data.} Recently, Reddit user, \texttt{Stuck\_In\_the\_Matrix}\footnote{https://www.reddit.com/user/Stuck\_In\_the\_Matrix/}, made available large data dumps that contain a majority of the content (posts and comments) generated on Reddit\footnote{http://couch.whatbox.ca:36975/reddit/}. The data dumps, collected using the Reddit API, are organized by month and year.  The data date back to 2006 and are regularly updated with new content. We use all comments from January 2006 through January 31, 2016 and expanded the dataset with each update. Each file corresponds to a month of Reddit data, and every line is a json object of a Reddit comment or post.

For our analysis, we identify three commonly targeted groups on Reddit --- African-American (black), plus-sized (plus) and women. For each of the target groups, we select the most active support and hate subreddits. To create our datasets, we extract all user comments in the selected subreddits from the data dumps described above, in October 2015. The details on the selected subreddits and the number of the extracted comments are provided in Table \ref{TAB:data_reddit}. 

\paragraph{Methods.} For each of the selected subreddits, we use labeled Latent Dirichlet Allocation (LLDA) to learn the topics that characterize them, against a baseline Reddit language.  This baseline is intended to push the LLDA to remove non-topical vocabulary from the two subreddit topics; it consists of a sample of 460,000 comments taken at random from the Reddit data scrape (none of the posts belonged to any of the subreddits of interest).  Prior to topic modeling, stop words, punctuation, URLs, and digits were stripped from the comments and for the purpose of balanced analysis, an equal number of comments was selected from the subreddit and the random sample. We use JGibbLDA for the topic inference \cite{phan2006jgibblda}.

\begin{table}[]
\centering
\begin{tabularx}{.49\textwidth}{XX|XX|XX}
\multicolumn{2}{c}{Black}                 & \multicolumn{2}{c}{Plus-size}           & \multicolumn{2}{c}{Female}             \\
\hline
Coon-   & racism    & FPH   & loseit    & TRP   & TwoXCr    \\
Town    &           &       &           &       &           \\
\hline
\hline
nigger               & \textbf{white}     & \textbf{weight}   & \textbf{weight}     & \textbf{women}  & \textbf{time}        \\
\textbf{white}       & racism             & \textbf{calorie}  & \textbf{calorie}    & \textbf{girl}   & \textbf{women}       \\
\textbf{black}       & \textbf{black}     & \textbf{time}     & \textbf{time}       & \textbf{time}   & \textbf{feel}        \\
shit                 & \textbf{racist}    & \textbf{work}     & \textbf{food}       & woman           & \textbf{work}        \\
\textbf{time}        & \textbf{race}      & \textbf{food}     & \textbf{eating}     & \textbf{shit}   & \textbf{year}        \\
fucking              & \textbf{time}      & \textbf{feel}     & \textbf{week}       & \textbf{work}   & \textbf{fuck}        \\
fuck                 & person             & \textbf{eating}   & \textbf{work}       & \textbf{year}   & \textbf{shit}        \\
\textbf{race}        & point              & \textbf{week}     & \textbf{feel}       & \textbf{life}   & weight               \\
year                 & feel               & \textbf{lose}     & \textbf{lose}       & \textbf{fuck}   & \textbf{fucking}     \\
hate                 & comment            & \textbf{year}     & \textbf{diet}       & guy             & person               \\
\textbf{racist}      & american           & women             & \textbf{body}       & point           & \textbf{life}        \\
live                 & post               & \textbf{diet}     & exercise            & friend          & \textbf{girl}        \\
work                 & issue              & \textbf{body}     & \textbf{goal}       & post            & love                 \\
jew                  & asian              & start             & loss                & \textbf{feel}   & pretty               \\
crime                & color              & \textbf{goal}     & \textbf{year}       & \textbf{fucking}& food                 \\
\hline
\multicolumn{2}{c}{Jaccard Index: 0.28} & \multicolumn{2}{c}{JI: 0.76}    & \multicolumn{2}{c}{JI: 0.50}
\end{tabularx}
\caption{Top discovered topics from support and hate subreddits for the three targets. The bold terms signify those that are present in both the hate and support vocabulary.}
\label{TAB:topic_overlap}
\end{table}

\paragraph{Results.} In Table \ref{TAB:topic_overlap}, we present the 15 most topical words from each subreddit.  The top terms in the topics are consistent with the target/support communities. For example, the term  ``women'' was ranked highly in subreddits that concern women (whether positively or negatively referenced) and ``weight" is the highest ranked topic for subreddits discussing plus-sized individuals and lifestyle. 

We observe a substantial overlap in vocabulary of hate and support subreddits, across all three target communities (see bold words in Table \ref{TAB:topic_overlap}).  While in the case of a black target group, we observe a Jaccard Index ({\it JI}) of 0.28, the overlap is higher in the case of female targets with {\it JI} at 0.50 and much higher for plus-size targets, with a {\it JI} of 0.76. 

The implication of this shared vocabulary is that while keywords can be used to detect text relevant to the target, they are not optimal for detecting targeted hateful speech. Shared vocabulary increases the likelihood of tagging content that is related to the target but not necessarily hateful, as hateful and increases false positives. We therefore require more robust training data.

\section{A community-driven model of hateful speech}

A key objective of our research is to avoid the issues associated with using manual annotation and keyword searches to produce training data for a classifier. As noted previously, sociological literature acknowledges that communities are formed by coherent linguistic practices and are defined, in part, by their linguistic identity \cite{gumperz2009speech}. Thus, the opportunity considered here is to leverage the linguistic practices of specific online communities to empirically define a particular kind of hateful speech. 

Since linguistic practices coincide with the identity of a community using them, we can define hateful speech as discourse practiced by communities who self-identify as hateful towards a target group. The members of the community contribute to the denigration of the target and, therefore, share a common linguistic identity. This allows us to develop a language model of hateful speech directly from the linguistic conventions of that community without requiring manual annotation of specific passages or keyword-based searches. This approach has a number of advantages over these practices.

First, a community-based definition removes the interpretive challenge involved in manual annotation. Membership in a self-organized community that is committed to denigration of a target group through the hatred of others is an observable attribute we can use to surface hateful speech events.

Second, unlike prior work, our method does not require a keyword list. We identify communities that conform to the linguistic identity of a self-organized hateful groups and use such communities to collect data. This data is used to learn the language model around the linguistic identity for detection. This removes any biases implicit in the construction of a keyword list (i.e., in the words included in or excluded from the list).

Third, a community-based definition provides a large volume of high quality, current, labeled data for training and then subsequent testing of classifiers. Such large datasets have traditionally been difficult to collect due to dependence on either manual annotation (annotation is slow and costly) or keyword searches (stringent keywords may turn up relatively few hits).

This approach generalizes to other online environments (such as Voat and other hateful speech-focused web forums) in which communities declare their identities, intentions, and organize their discussions.  Any online (or, even, offline) communication forum in which all participants gather for the understood purpose of degrading a target group constitutes a valid source of training data.

In the following subsections, this approach is validated through three analyses.  First, we demonstrate that the hate speech communities identified actually employ distinct linguistic practices: we show that our method can reliably distinguish content of a hateful speech community from the rest of Reddit.  We also show that our approach substantially outperforms systems built on data collected through keywords.

Second, we show that our approach is sensitive to the linguistic differences between the language of hateful and support communities. This task is notably difficult given the results we reported above, showing that such communities share many high-frequency words.

Finally, we use our Reddit-trained classifier to detect hateful speech on other (non-Reddit) platforms: on Voat and hateful speech web forums (websites devoted to discussion threads attacking or denigrating a target community). For both, we find that our method performs better than a keyword-based baseline.

\subsection{Data collection}
\noindent {\bf Reddit.} We use Reddit as the primary source for the hateful communities and leverage the linguistic practices of these communities to empirically define and develop language models for target-specific hateful speech. In all three of our studies, we focus on the aforementioned three target groups: black people, plus-sized individuals, and women. For each, we select the most active hateful and support subreddits and collect all the publicly available comments present in the data dumps provided by \texttt{Stuck\_In\_the\_Matrix}. The details on the dataset are provided in Table \ref{TAB:data_reddit}. We also collect a random sample of 460,000 Reddit comments to serve as negative examples.

\noindent {\bf Voat.} Voat, a content aggregator similar to Reddit, also hosts active discussion communities, called \textit{subverses}, few of which identify as hateful. We select Voat because of its similarity to our original source\footnote{http://thenextweb.com/insider/2015/07/09/what-is-voat-the-site-reddit-users-are-flocking-to/}. Since the two websites cater to a similar user-base, the generated linguistic identities should be similar in sub-communities with similar themes. Therefore, the language model of hateful communities on Reddit should match, to an extent, with the language model of similar hateful communities on Voat.

For the three target groups, we identify hateful subverses --- \texttt{v/CoonTown}, \texttt{v/fatpeoplehate} and \texttt{v/TheRedPill} --- sub-communities that share their name with their counterparts on Reddit and target blacks, plus-size individuals, and women, respectively. In the absence of an API, we use web-scraping libraries to retrieve all publicly available comments posted to the selected subverses between July 2015 and January 2016. We also collect a set of 50,000 comments (from the same time period) from a random sample of subverses to serve as negative examples (Table \ref{TAB:data_other_platform}).

\noindent {\bf Web forums.} We also use stand-alone web forums that are dedicated to expressing hate or contempt for the target communities. These web forums are social platforms that provide their users with discussion boards, where users can create threads under predefined topics and other users can then add comments in these threads. We, therefore, select web forums for their discussion-based communities and user-generated content. Again, due to the lack of APIs, we use, as data, comments that were collected by web-scraping libraries from numerous threads of their discussion boards during October 2015. 

For the black target group, we use \texttt{Shitskin.com}: our dataset consists of 3,160 comments posted to 558 threads from three of website's boards: ``Primal Instinct'', ``Crackin the whip!'' and ``Underground Railroad.'' For the female target group, we use \texttt{mgtowhq.com}: this dataset consists of 20688 comments posted to 4,597 threads from the ``MGTOW General Discussion" board. Finally, as a source of negative examples, we use the ``random'' discussion board on \texttt{topix.com}: this dataset consists of nearly 21,000 comments from 2458 threads. To our knowledge, no large fat-shaming forum exists, thus we do not include this target group in this phase of the study (Table \ref{TAB:data_other_platform}).  All comments have posting times between July 2015 and January 2016.

\begin{table}[]
\centering
 
\begin{tabularx}{.5\textwidth}{llrlr}
Target      & Subverse    & Voat      & Website  & Comments \\
\hline
\hline
Black       & CoonTown    & 3358      & shitskin & 3160   \\
Plus  & fatpeoplehate   & 31717     & -        &         \\
Female      & TheRedPill     & 478       & mgtowhq  & 20688  \\
\end{tabularx}
\caption{Target-relevant hateful comments collected from Voat subverses and web forums.}
\label{TAB:data_other_platform}
\end{table}

\begin{table*}[]
\centering
\begin{tabularx}{\textwidth}{l|XXX|XXX|XXX|XXX|XXX}

\multicolumn{10}{l}{(a) Assessing the distinct nature of language emerging from hate groups.}\\
\hline
Target & \multicolumn{3}{c}{Accuracy}                & \multicolumn{3}{c}{Precision} & \multicolumn{3}{c}{Recall} & \multicolumn{3}{c}{F1-Score} & \multicolumn{3}{c}{Cohen's $\kappa$} \\
       & NB       & SVM       & LR     & NB       & SVM  & LR   & NB   & SVM  & LR   & NB   & SVM  & LR & NB   & SVM  & LR   \\
\hline
\hline
Black  & 0.79 & 0.81 & 0.81 & 0.78 & 0.84 & 0.87 & 0.82 & 0.74 & 0.73 & 0.8  & 0.79 & 0.79 & 0.58 & 0.61 & 0.61 \\
Plus   & 0.78 & 0.78 & 0.79 & 0.78 & 0.81 & 0.82 & 0.79 & 0.75 & 0.73 & 0.78 & 0.78 & 0.77 & 0.56 & 0.57 & 0.57 \\
Female & 0.77 & 0.8  & 0.81 & 0.71 & 0.81 & 0.84 & 0.9  & 0.77 & 0.75 & 0.79 & 0.79 & 0.79 & 0.55 & 0.6  & 0.61 \\
\hline
\multicolumn{10}{l}{}\\
\multicolumn{10}{l}{(b) Assessing sensitivity between the language of hate and support groups.}\\
\hline
\hline
Black  & 0.8  & 0.79 & 0.79 & 0.8  & 0.8  & 0.78 & 0.85 & 0.82 & 0.86 & 0.82 & 0.81 & 0.82 & 0.57 & 0.56 & 0.55 \\
Plus   & 0.83 & 0.85 & 0.85 & 0.85 & 0.84 & 0.84 & 0.79 & 0.86 & 0.86 & 0.82 & 0.85 & 0.85 & 0.66 & 0.69 & 0.7  \\
Female & 0.79 & 0.78 & 0.78 & 0.78 & 0.79 & 0.8  & 0.79 & 0.77 & 0.77 & 0.79 & 0.78 & 0.78 & 0.57 & 0.56 & 0.57
\end{tabularx}
\caption{The performance of the three classification algorithms across the three target groups, with a 10 fold cross-validation. (a) Hateful comments are classified against random comments. (b) Hateful comments are classified against comments from support communities. In both cases, the classifier is able to distinguish hate speech from negative cases. (NB: Naive Bayes, SVM: Support Vector Machines, LR: Logistic Regression) }
\label{TAB:results_inreddit}
\end{table*}

\begin{table*}[]
\centering
\begin{tabularx}{\textwidth}{l|XXX|XXX|XXX|XXX|XXX}

\multicolumn{10}{l}{(a) Baseline performance over Reddit data.}\\
\hline
Target & \multicolumn{3}{c}{Accuracy}                & \multicolumn{3}{c}{Precision} & \multicolumn{3}{c}{Recall} & \multicolumn{3}{c}{F1-Score} & \multicolumn{3}{c}{Cohen's $\kappa$} \\
       & LDA & $\chi^2$I & $\chi^2$II & LDA & $\chi^2$I & $\chi^2$II & LDA & $\chi^2$I & $\chi^2$II & LDA & $\chi^2$I & $\chi^2$II & LDA & $\chi^2$I & $\chi^2$II \\
\hline
\hline
Black  & 0.59 & 0.63 & 0.57 & 0.61 & 0.71 & 0.62 & 0.52 & 0.44 & 0.4  & 0.56 & 0.54 & 0.48 & 0.18 & 0.26 & 0.15 \\
Plus   & 0.53 & 0.57 & 0.53 & 0.54 & 0.6  & 0.55 & 0.35 & 0.4  & 0.34 & 0.42 & 0.48 & 0.42 & 0.06 & 0.14 & 0.06 \\
Female & 0.68 & 0.7  & 0.7  & 0.65 & 0.69 & 0.74 & 0.71 & 0.71 & 0.6  & 0.68 & 0.7  & 0.66 & 0.35 & 0.40 & 0.4  \\
\hline
\multicolumn{10}{l}{}\\
\multicolumn{10}{l}{(b) Baseline performance over Voat data.}\\
\hline
\hline
Black  & 0.62 & 0.63 & 0.62 & 0.65 & 0.73 & 0.68 & 0.48 & 0.4  & 0.4  & 0.55 & 0.51 & 0.51 & 0.24 & 0.26 & 0.23 \\
Plus   & 0.56 & 0.6  & 0.57 & 0.58 & 0.65 & 0.61 & 0.35 & 0.4  & 0.36 & 0.43 & 0.5  & 0.45 & 0.11 & 0.2  & 0.14 \\
Female & 0.67 & 0.69 & 0.67 & 0.68 & 0.71 & 0.74 & 0.63 & 0.63 & 0.5  & 0.65 & 0.67 & 0.6  & 0.35 & 0.38 & 0.34 \\
\hline
\multicolumn{10}{l}{}\\
\multicolumn{10}{l}{(c) Baseline performance over web forum data.}\\
\hline
\hline
Black  & 0.66 & 0.62 & 0.57 & 0.72 & 0.77 & 0.67 & 0.53 & 0.35 & 0.31 & 0.61 & 0.48 & 0.42 & 0.32 & 0.24 & 0.15 \\
Female & 0.78 & 0.79 & 0.77 & 0.81 & 0.83 & 0.87 & 0.75 & 0.74 & 0.64 & 0.78 & 0.78 & 0.74 & 0.56 & 0.58 & 0.54
\end{tabularx}
\caption{We calculate the baseline performance on multiple platforms with three keyword-generating methods: LDA, $\chi^2$I and $\chi^2$II. Classification was done using logistic regression.}
\label{TAB:results_baseline}
\end{table*}

\begin{table}[]
\centering
\begin{tabularx}{.48\textwidth}{Xccccc}
Target    & Acc  & Pre  & Rec  & F1 & $\kappa$\\
\textit{Voat}      &      &      &      &      &\\
\hline
\hline
Black  & 0.82 & 0.87 & 0.74 & 0.80  & 0.64 \\
Plus   & 0.81 & 0.85 & 0.74 & 0.79 & 0.62 \\
Female & 0.74 & 0.76 & 0.71 & 0.73 & 0.49 \\
\textit{Websites}   &      &      &      &     & \\
\hline
\hline
Black  & 0.82 & 0.87 & 0.77 & 0.82 & 0.65 \\
Female & 0.77 & 0.83 & 0.69 & 0.75 & 0.54
\end{tabularx}
\caption{For our targets, we collect comments from hateful communities on Voat and web forums and test the performance of language models learned from Reddit communities.}
\label{TAB:results_crossplatform}
\end{table}

\begin{table}[]
\centering
\begin{tabularx}{.48\textwidth}{llccccc}
Training   & Testing    & Acc  & Pre  & Rec  & F1 & $\kappa$\\
\hline
\hline
CT      & FPH      & 0.58 & 0.72 & 0.26 & 0.38 & 0.15 \\
CT      & TRP      & 0.55 & 0.6  & 0.22 & 0.32 & 0.08 \\
FPH     & TRP      & 0.58 & 0.65 & 0.3  & 0.41 & 0.15 \\
FPH     & CT       & 0.54 & 0.61 & 0.23 & 0.34 & 0.08 \\
TRP     & CT       & 0.51 & 0.53 & 0.28 & 0.36 & 0.03 \\
TRP     & FPH      & 0.6  & 0.65 & 0.41 & 0.51 & 0.19
\end{tabularx}
\caption{We test the performance of classification systems built on data that belongs to a target community differnt than the one we test on. (CT: CoonTown) }
\label{TAB:results_Crosstarget}
\end{table}

\subsection{Methods}

Before the classification process, we preprocess all the data by eliminating URLs, stopwords, numerals and punctuations. We further lowercase the text and remove platform-relevant noise (e.g., comments from house keeping bots on Reddit like AutoModerator). The text is finally tokenized and used as input for the classification pipeline.

We use multiple machine learning algorithms to generate the language models of hateful communities. From the analysis of the prior work, we identify the commonly-used algorithms and employ them in our analysis. Specifically, we use naive Bayes (NB), support vector machines (SVM) and logistic regression (LR). We do this in order to assess the merits of our insight into using community-defined data collection.

The algorithms take as input, tokenized and preprocessed arrays of user comments along with the label of the community they belong to. We use a sparse representation of unigrams with \textit{tfidf} weights as our feature set. In future investigation, we would like to add part of speech tags and sentiment score as features. 

For performance evaluation, we use the standard measures: accuracy, precision, recall and F1-Score. We also use Cohen's $\kappa$ as a measure of agreement between the observed and expected labels. $\kappa$ helps in evaluating the prediction performance of classifiers by taking in account any chance agreement between the labels.

\paragraph{Baseline comparison.} Our aim is to assess the impact of using community-based text compared with keyword-based text as training data. Due to space limitations, here we report only a logistic regression classifier trained on keyword-collected data (SVM and NB showed comparable performance).

The specific keywords used are generated from the comments collected from hateful Reddit communities.  For a given target group, we generate three sets of keywords for each: (1) keywords generated between hate subreddits and a random sample of Reddit comments using LLDA, as in Section 3, (2) keywords generated between hate subreddits and a random sample of Reddit comments using $\chi^2$ weights ($\chi^2$I), and (3) keywords generated between hate and support subreddits using $\chi^2$ weights ($\chi^2$II). To generate the training datasets, we use the top 30 keywords and from a separate random sample of Reddit comments, collect samples that contain at least one of the keywords as positive samples and samples that contain no keywords as negative samples. For each keyword type and each target, we aggregate 50,000 positive and 50,000 negative samples for training.

\subsection{Results and Discussion}

\paragraph{Community language vs. hateful speech.} It may seem that, by comparing classifiers on the task of detecting hateful community posts, we are equating language produced by a hateful community with hateful language.  Certainly, they are not always the same. Some content is likely non-hateful chatter. One alternative for excluding such noise is manual coding of testing data.  Given the existing issues with such labeled data, we avoid such manual labeling. Furthermore, a comparison of the two approaches is not fair due to the associated trade-offs. The community definition, as mentioned, relies on the assumption that all the content in a hateful community is hateful, which might not always be true. However, such an assumption allows us to generate large training datasets with relative ease. We therefore allow the presence of some noise in the training data for ease of training data generation and favouring recall. On the other hand, manual annotation promises less noisy datasets at the expense of time and resources, which limits the size of training datasets. It would be very laborious to produce datasets as large as those generated with our community approach. Also, since manual annotation relies heavily on personal perception, it can also introduce noise in the datasets. In other words, manual annotation does not allow us to generate large training sets, and also cannot provide completely noise-free data.

Another option, however, is to focus on the precision ($\frac{TP}{TP+FP}$) of the classifier.  Precision indicates the classifier's ability to identify only content from the hateful community.  The construction of the test datasets is such that hateful speech should only exist in the hateful community posts.  Thus, a method that detects hateful content should strongly favor including only content from hateful communities --- yielding high precision.  Crucially, in the discussions that follow, we find that a community-based classifier demonstrates much higher precision than keyword-based methods.  Thus, by either measure (F1 or precision), our community-based classifier outperforms the baselines.

\paragraph{Hateful groups have distinct linguistic signatures.} In Table \ref{TAB:results_inreddit}(a), we see the performance of the three classifiers when classifying a balanced corpus of hateful posts and randomly selected (non-hateful speech) Reddit posts with 10-fold cross validation. The dataset consists of all the comments collected from the relevant hate subreddit (Table \ref{TAB:data_reddit}) as positive samples and an equal number of random Reddit comments as negative samples. We observe the three classifiers perform almost identically. Naive Bayes slightly outperforms others on Recall and F1-score, while Logistic Regression is a slightly better performer on the other metrics. Also, the performance of the classifiers is consistent across the three target groups. Analysis of $\kappa$ suggests that observed labels after the classification process are in moderate to substantial agreement with the expected labels.

\noindent {\bf Comparison to baseline.}  In all cases considered, a classifier trained on community-based data outperforms a keyword-based classifier. Notably, the keyword-based classifier for the women-target group performed best, suggesting that hateful community language associated with the keywords used for collection are more representative of hateful speech (compared to other communities).  

From a precision perspective, we find that the community-based classifier outperforms the baselines by between $10\%$ and $20\%$, indicating that the community-based classifier is including far fewer incorrect cases of hateful speech (false positives). When we look at the true positive posts that have been detected exclusively by the community-based classifier (i.e., that the keyword-based approach missed), we find many that are clearly hateful, but in ways that do not use specialized slurs.  Several examples from the \texttt{CoonTown} subreddit:

\begin{enumerate}
\item ``I don't see the problem here. Animals attack other animals all the time.''
\item ``Oy vey my grandparents vuz gassed ven dey vaz six years old!''
\item ``DNA is rayciss, or didn't you know?''
 \item ``Are they going to burn their own town again? Yawn.''
\end{enumerate}

These examples characterize different (and important) ways in which speech can be hateful without using words that typically operate, largely independent of context, as slurs.  In Example 1, African-Americans are described as animals, employing a word that is not usually a slur, to denigrate them. In Example 2, historical context (the gas chambers in Nazi concentration camps), culturally stereotyped language (``Oy vey''), and spelling to imitate an accent (``ven dey vaz'') are successfully used to express contempt and hatred, without any slur or even any word that, like 'animals' in the first example, is sometimes pressed into service as a slur.  The third example, like the second, parodies an accent, and here it is notable that while ``racist'' might be a keyword use for collection, it's unlikely that ``rayciss'' would be used.  Finally Example 4 achieves its effect by attacking a group through an implication of stereotyped action without even actually naming them at all (as opposed to Example 1, in which the targets were called ``animals'').

\paragraph{Community-based approach is sensitive to the linguistic differences of hate and support communities.} In Section 3, we showed that hateful and support communities for a target group have a shared vocabulary: the two communities often engage in discourse on similar topics, albeit with quite different intent. Since the shared keywords are not effective in the discrimination process, recognizing the distinction between hate and support communities can be challenging. We set up a classification task for identifying comments from support and hate communities, carried out with a 10-fold cross-validation. The performance of the task is presented in Table \ref{TAB:results_inreddit}(b). We observe that this performance is close to the performance of our system against a random collection of Reddit comments (Table \ref{TAB:results_inreddit}(a)). Therefore, even with shared vocabulary, our system is sensitive to the distinction in linguistic characteristics of hateful and support communities for the same target.

\paragraph{Community-trained systems can be deployed on other platforms.} Often training data for hateful language classification can be hard to obtain on specific platforms.  For this reason, methods that work across platforms (trained on one platform, applied on another platform) present significant advantages.

For the analysis, we continue with the same three target groups and train our language model, using logistic regression, with comments from relevant Reddit communities and then test it on data we collected from other platforms. The performance of the system, (Table  \ref{TAB:results_crossplatform}), is very similar to the results we obtain when testing on Reddit (Table \ref{TAB:results_inreddit}(a)). This said, we must be careful not to overstate our method's generalizability.  While, certainly, the degree of generalizability observed is noteworthy (particularly given past work), these platforms all feature similar posting conventions: posts are not length restricted, are made within well defined discussion threads, and have a clear textual context.  Our method will likely perform well on any such forum-based system.  Platforms, which involve quite different conventions, particularly those that are predominantly populated by short-text posts (e.g., Twitter and Facebook), will likely involve additional work. Nonetheless, we do believe that the community-based approach presents opportunities for these other platforms as well.

\paragraph{Hateful classifiers are not target-independent.} Hateful conversations are thematic and major topics discovered from conversations are target related  (Table \ref{TAB:topic_overlap}). Not surprisingly, our system performs poorly when tested across targets. We train the classifier on one target and test it on another. The results (see Table \ref{TAB:results_Crosstarget}) provide a strong indication that hateful speech classification systems require target-relevant training.

\paragraph{Detailed Error Analysis.} In order to better understand the performance of our system, we manually inspect a set of erroneously classified posts from the \textit{coontown} training/testing dataset.  We characterize the kinds of issues we observe and discuss them here.

\noindent {\it Type I errors.} These posts arise when non-hate group posts are labeled as hate-group posts.  Notably, we observe that some of these errors are actually racist comments that originated from other communities in Reddit.

\begin{enumerate}
\item ``well jeez if u pit a nigger against a cunt what do u expect"
\item ``Triskaid is a fucking nigger."
\end{enumerate}

In both of the cases the comments were in fact racist and were therefore correctly labeled.  This, of course, points out a potential (though, we would argue minor) weakness of our approach, which is that hate groups are not the {\it only} source of hateful language --- simply the most high-density source. 

More frequently, Type I errors featured non-racist comments which had been mislabeled. This is likely due to the fact that not all content in a hateful community is hateful: some is simply off-topic banter among community members.  This adds noise during the training phase which manifests as classification errors.  While certainly an issue, given the dramatic improvement in overall classification performance, we consider this an acceptable trade off at this stage in the research. Future work should consider ways of focusing training data further on the distinctly hateful content produced by these communities.

\noindent {\it Type II errors.} In most cases where hateful-speech community posts were incorrectly labeled as non-hateful, we primarily find that these were, in fact, non-racist posts that were made to the hateful subreddit.  Here are a few examples:

\begin{enumerate}
\item ``and you're a pale virgin with a vitamin d deficiency."
\item ``Whats the deal with you 2? And besides, we're all on the same side here.."
\item ``IP bans do literally nothing, it only takes a moment to change it."
\item ``I can't believe Digg is still up. I can't believe Reddit is still up."
\end{enumerate}

Posts like these constitute noise, in terms of our community-based definition of hateful speech, discussed above. Nonetheless, our system was able to correctly identify them as non-hateful. Taken together with the Type I errors, it appears that the noise implicit in our community-definition of hateful speech yields a modest increase in Type I error, but can somewhat be removed by the classifier in the form of Type II errors (which are not, in fact, errors).

A very small number of other Type II errors are examples of hateful speech, but that target a community other than blacks (in the cases we saw, primarily Jews):

\begin{enumerate}
\item ``Peace and harmony? Yeah that's why they stole that land (now kikeriel) and killed the civilians that lived there before. Did I mention they STILL kill the Palestinians to this day and cover it up? Fuck them."
\item ``quit kissing kikeass"
\item ``You sound like a jew. In a system ruled by money, money can buy anything. Everything is capitalisms fault. But I get why you'd support capitalism since your ``people" invented the whole shebang"
\item  ``Losing weight isn't even hard, stop eating like a fucking landwhale, drink lots of water and move your fatass"
\end{enumerate}

Although these comments are hateful, since they are not directed at black people, the system is technically performing according to specification. 

Our system missed some cases of obvious racism, such as the following examples. However, such cases constitute only a small fraction of the comments in Type II error.

\begin{enumerate}
\item ``Ok Korea - you know your duty in the impending `blackification' of the globe? I know where I stand"
\item ``Black people are terrible. " 
\item ``Pretty soon we will need a dedicated sub for black-on-senior sexual assaults."
\item ``Who is the target audience? I would think black literacy levels would prevent ``nig lit" from ever being a viable book market."
\end{enumerate}

Overall, our analysis of Type II errors indicated that the vast majority of mislabeled comments are not racist and are, therefore, correctly labeled. This suggests that the actual performance of our method is likely higher than what we report.

\paragraph{Imbalanced Datasets} We use balanced datasets for our analysis. Since this assumption may or may not hold for different data sources, we perform some initial analysis on imbalanced datasets. As the actual composition of data sources can be variable, we generate testing sets with the ratio of hateful content to non-hateful content at 1:10, 1:100, 1:1000. Our preliminary results are similar to the performance on a balanced test set. These results are encouraging but require further analysis. We hope to overcome the challenges of dataset-shift due to mismatch in the composition of testing and training datasets in future work.

\section{Conclusion}

The presence of hateful speech on online platforms is a growing problem with a need for robust and scalable solutions. In this work, we investigated the limitations of keyword-based methods and introduced a community-based training method as an alternative. Our work makes two key contributions.

First, we highlight two major mechanisms that hurt the performance of keyword-based methods.  The shared vocabulary between hateful and support communities causes training positive examples to contain non-hateful content.  Also, because keyword lists focus on more widely known slurs, these lists miss many instances of hateful speech that use less common or more nuanced constructions to express hatred all too clearly.

Our second contribution is the idea of using self-identified hateful communities as training data for hateful speech classifiers.  This approach both involves far less effort in collecting training data and also produces superior classifiers.

The promising results obtained in this study suggest several opportunities for future work. Foremost is the extension of this approach to other non-forum-based platforms.  Twitter and Facebook, for example, are heavily used platforms which mainly feature short-text messages.  Such content presents unique challenges that will require new or modified approaches. Another direction involves looking at other high-signal features (syntax, n-grams, and sentiment scores).

In these and other initiatives, we believe that community-based data may play an essential role in producing both better detectors of hateful speech, and a richer understanding of the underlying phenomenon.

\section{Bibliographical References}
\label{main:ref}

\bibliographystyle{lrec2016}
\bibliography{tacos}

\end{document}